\documentclass{article}
\usepackage{arxiv}
\usepackage{graphicx}
\usepackage{hyperref}

\usepackage{xcolor}
\usepackage[font=small]{caption}
\usepackage{float}
\usepackage{subcaption}
\usepackage{mwe}
\usepackage{amsmath}
\usepackage{bm}
\usepackage{array}
\usepackage{multirow}
\usepackage{longtable}
\usepackage{adjustbox}
\usepackage{tcolorbox}

\title{Reasoning and Sampling-Augmented MCQ Difficulty Prediction via LLMs}

\author{{Wanyong Feng} \\
	University of Massachusetts Amherst \\
        Amherst, MA, USA \\
	\texttt{wanyongfeng@umass.edu}
	\And
        {Peter Tran} \\
	University of Massachusetts Amherst \\
        Amherst, MA, USA \\
	\texttt{petertran@umass.edu}
	\And
        {Stephen Sireci} \\
	University of Massachusetts Amherst \\
        Amherst, MA, USA \\
	\texttt{sireci@acad.umass.edu}
	\And
        {Andrew Lan} \\
	University of Massachusetts Amherst \\
        Amherst, MA, USA \\
	\texttt{andrewlan@cs.umass.edu}
}

\renewcommand{\shorttitle}{\textit{arXiv} Template}

\hypersetup{
pdftitle={A template for the arxiv style},
pdfsubject={q-bio.NC, q-bio.QM},
pdfauthor={David S.~Hippocampus, Elias D.~Striatum},
pdfkeywords={First keyword, Second keyword, More},
}

\begin{document}
\maketitle              
\begin{abstract}
The difficulty of multiple-choice questions (MCQs) is a crucial factor for educational assessments. Predicting MCQ difficulty is challenging since it requires understanding both the complexity of reaching the correct option and the plausibility of distractors, i.e., incorrect options. In this paper, we propose a novel, two-stage method to predict the difficulty of MCQs. First, to better estimate the complexity of each MCQ, we use large language models (LLMs) to augment the reasoning steps required to reach each option. We use not just the MCQ itself but also these reasoning steps as input to predict the difficulty. Second, to capture the plausibility of distractors, we sample knowledge levels from a distribution to account for variation among students responding to the MCQ. This setup, inspired by item response theory (IRT), enable us to estimate the likelihood of students selecting each (both correct and incorrect) option. We align these predictions with their ground truth values, using a Kullback-Leibler (KL) divergence-based regularization objective, and use estimated likelihoods to predict MCQ difficulty. We evaluate our method on two real-world \emph{math} MCQ and response datasets with ground truth difficulty values estimated using IRT. Experimental results show that our method outperforms all baselines, up to a 28.3\% reduction in mean squared error and a 34.6\% improvement in the coefficient of determination. We also qualitatively discuss how our novel method results in higher accuracy in predicting MCQ difficulty. 


\end{abstract}

\section{Introduction}
Assessing student knowledge and skill levels is important in education. Effective assessments should evaluate whether students have acquired sufficient knowledge and necessary skills. Since assessments are composed of individual questions (items), ensuring the quality and fairness of these questions is crucial. According to the Standards for Educational and Psychological Testing \cite{eignor2013standards}, various factors must be considered to uphold the quality and validity of each question. Among these, one of the most critical factors is question difficulty, a numerical measure that reflects how challenging the question is to answer. Question difficulty plays a vital role in various intelligent educational applications such as Computerized Adaptive Testing (CAT) \cite{feng2023balancing,ref1,ref2} and Automatic Question Generation (AQG) \cite{jiao2023automatic,kurdi2020systematic,shimmei2022automatic}. In CAT, question difficulty is used to adaptively select questions that accurately assess different student knowledge levels while minimizing the exam length. In AQG, question difficulty is used to evaluate the quality of generated questions, ensuring that the generated questions has appropriate difficulty levels.

Among various question formats, multiple-choice questions (MCQs) are one of the most widely used in educational assessments. MCQs are popular due to their efficiency, since they allow for quick administration and grading, making them particularly suitable for large-scale assessments \cite{hwang2024towards,thomas2024does}. 
A MCQ typically consists of a structured format: the stem, which provides the question setup and context, followed by a question to be answered. The options include the correct option, referred to as the key, and several incorrect options, referred to as distractors. See figure~\ref{fig:training_process} for a specific example. 

Predicting MCQ difficulty is challenging, since it requires assessing both the complexity of reaching the key and the plausibility of the distractors. Existing methods primarily focus on extracting various features from MCQ to predict difficulty \cite{benedetto2020introducing,yaneva2019predicting}. More recent methods incorporate Large Language Model (LLM)s’ responses, which approximate student responses, to MCQ as input for feature extraction \cite{duenas2024upn,rogoz2024unibucllm}. However, we argue that these methods have two key limitations. From the student perspective, incorporating their response data is crucial since these data provide an understanding of how students interact with the MCQ. More importantly, a comprehensive set of student response data is essential to ensure that the predicted difficulty accounts for students with varying knowledge levels. Without such diversity, such as using LLMs to approximate limited number of students, methods may overfit a specific group, leading to biased predictions. From the MCQ perspective, relying solely on MCQ content makes it difficult for methods, even LLMs, to extract features that can represent the complexity of reaching the key and the plausibility of distractors accurately.

\subsection{Contributions}
In this paper, we propose a novel method that addresses the two limitations mentioned above to predict MCQ difficulty using LLMs. We note that our method needs the actual student response data to MCQs, i.e., which option each student selects. Our method treats MCQ difficulty prediction as a four-step process: 
\begin{itemize}
\item We use \textit{GPT-4o} to generate reasoning steps required to reach the key, and feedback messages explaining the potential errors that might lead students to select each distractor.
\item We use \textit{Longformer}, an encoder-only based LLM, to extract the latent feature of each option. We use the stem combined with key or distractor paired with its corresponding reasoning steps or feedback messages as input. These reasoning steps and feedback messages serve as auxiliary information, enabling \textit{Longformer} to better capture the complexity of reaching the key and the plausibility of the distractors. 
\item We sample a comprehensive set of student knowledge levels from the standard multivariate normal distribution inspired by the Item Response Theory (IRT). We use a Multi-Layer Perceptron (MLP) to predict the likelihood that these students, on average, will select each option of a MCQ.
\item We use another MLP to predict the MCQ difficulty using the predicted student selection likelihood of all options. 
\end{itemize}
The intuition behind our method is that the likelihood of average students selecting options is closely related to MCQ difficulty. For instance, a high likelihood of selecting the key suggests that the MCQ is relatively easy, whereas a low likelihood indicates higher difficulty. Additionally, when the likelihood of selecting all options is more evenly distributed, the MCQ is more challenging since distractors are plausible. We ensure the accuracy of predicted average student selection likelihood distribution from two perspectives. From the student perspective, to account for students with varying knowledge levels, we build on IRT, in which student knowledge levels are conceptualized as samples drawn from a standard normal distribution. We sample a comprehensive set of student knowledge levels from a multivariate standard normal distribution. From the MCQ perspective, reasoning steps and feedback messages provide rich contextual information that enhances the \textit{Longformer}’s ability to understand the complexity of reaching the key and the plausibility of the distractors. During training, to further improve the accuracy of the predicted distribution, we align the predicted distribution with the ground truth distribution, which is derived from actual student response data. This alignment is achieved by minimizing the Kullback-Leibler (KL) divergence \cite{kl} between the two distributions.
We evaluate our method on two datasets containing math MCQs, student response data, and IRT-based ground truth difficulty. Through a series of quantitative experiments, our results demonstrate that our proposed method outperforms baselines, often by a significant margin. Notably, on one dataset, our method achieves a substantial Mean Squared Error reduction of 28.3\% and a Coefficient of
Determination improvement of 34.6\% compared to the best baseline. We also conduct qualitative experiments and discuss the potential reasons of why our method can achieve more accurate MCQ difficulty prediction.

\section{Related Work}
Given the importance of the MCQ difficulty, extensive research has focused on developing accurate MCQ difficulty prediction methods. See \cite{review1,review2,review3} for more detailed reviews. Traditional methods rely primarily on two approaches: experts’ judgments, leveraging their domain knowledge and experience \cite{choi2020predicting}, and pre-testing \cite{rust2014modern}. However, such methods are known to be time-consuming, subjective, and often inconsistent \cite{benedetto2020r2de,hsu2018automated}. 

To address these limitations, the rise of Artificial Intelligence (AI) has introduced alternative solutions. Natural Language Processing (NLP) techniques have been widely used for automatic MCQ difficulty prediction. These methods focus on extracting textual features and using these features to train predictive models. 
Commonly used surface-level textual features include word length, the number of sentences/clauses \cite{pandarova2019predicting}, and the frequency of academic words \cite{loukina2016textual}. More sophisticated textual features include Flesch-Kinkaid readability score \cite{kincaid1975derivation}, which several researchers note that it is a weak predictor \cite{yaneva2019predicting,benedetto2020introducing}, and Term-frequency-inverse document frequency (TF-IDF) \cite{salton1983modern}. Word2Vec embeddings \cite{mikolov2013efficient} are extensively used in MCQ difficulty prediction. For example, \cite{yaneva2019predicting} uses Word2Vec embeddings alongside linguistic and psycholinguistic features to predict the difficulty of medical MCQs. From their ablation study results, they find that word embeddings are the most predictive features. Moreover, with the advancement of neural networks, the Bidirectional Long Short-Term Memory (BiLSTM) network is commonly used to capture latent features of the MCQ components for MCQ difficulty prediction. For example, \cite{qiu2019question} uses BiLSTM to encode MCQ components, using the latent features to predict confusion difficulty and recall difficulty. These two predictions are then combined for the final difficulty prediction.


More recently, with the development of LLMs, LLMs are used to capture latent features of MCQs, leading to improved difficulty prediction accuracy \cite{benedetto2021application,xue2020predicting,zhou2020multi}. Especially, based on the findings of \cite{benedetto2021application}, LLM-extracted features outperform those used in previous works. Furthermore, they show that pretraining the LLM on external textual sources relevant to the MCQ topic before finetuning for difficulty prediction further enhances performance. \cite{loginova2021towards} proposes a new method to predict difficulty using a trained question-answering (QA) LLM. Specifically, the QA LLM first predicts the likelihood of selecting each option. They then calculate the variance of this likelihood distribution as the predicted difficulty.

\section{Methodology}
In this section, we formally define the MCQ difficulty prediction task, detail our method and the intuition behind it, and the training process.
\subsection{Task Definition}
Given a dataset $D$ that contains $N$ MCQs, We define the $i$-th MCQ $Q_i$ as consisting of a set of components: 
\begin{equation*}
Q_i = \{s_i, k_i, n_{k_i}, d_{i_1}, n_{d_{i_1}}, d_{i_2}, n_{d_{i_2}}, d_{i_3}, n_{d_{i_3}}\}.
\end{equation*}
$Q_i$ contains a stem $s_i$, a key $k_i$, the number of students who select the key $n_{k_i}$, (without loss of generality, three) distractors $d_{i_1}$, $d_{i_2}$, and $d_{i_3}$, with their respective numbers of student selections denoted as $n_{d_{i_1}}$, $n_{d_{i_2}}$, and $n_{d_{i_3}}$. We formulate the MCQ difficulty prediction task as a regression problem, where the goal is to train a model that takes all components of $Q_i$ as input and outputs a real-valued difficulty, which can be obtained from e.g., IRT.

\subsection{Method Modules}
Our method has four modules, which we detail below; Figure~\ref{fig:training_process} illustrates the entire pipeline for MCQ difficulty prediction.

\textbf{Reasoning/Feedback Generation Module}
The first module, $\theta_{gen}$, takes the stem and a single option as input. If the option is the key, it generates the reasoning steps required to reach the key. If the option is a distractor, it generates a feedback message explaining the potential error a student might make in selecting that distractor. We simply prompt GPT-4o, a powerful proprietary LLM that
possesses strong mathematical reasoning abilities, for this module. We prompt \textit{GPT-4o} with detailed instructions and 4 demonstrations to guide it to generate reasonable reasoning steps and feedback messages. Given the limited space, we provide one demonstration MCQ with only the key and its corresponding reasoning steps:
\begin{tcolorbox}
\texttt{Question: The sum of three consecutive even numbers is 48. What is the middle value?} \newline
\texttt{Correct Answer: 16} \newline
\texttt{Reasoning steps for the correct answer: To find the middle value of three consecutive even numbers that sum up to 48, let's represent the three consecutive even numbers as x, x+2, and x+4. Here’s the step-by-step solution: 1. Add the three numbers together: x + (x+2) + (x+4) = 48; 2. Combine like terms: 3x + 6 = 48; 3. Subtract 6 from both sides to isolate the term with x: 3x = 42; 4. Divide both sides by 3 to solve for x: x = 14 So, the three consecutive even numbers are 14, 16, and 18. The middle number is 16.}
\end{tcolorbox}
We denote the generated reasoning steps and feedback messages as $r_i$ and $f_{i_\ell}$ respectively: 
\begin{equation}
r_i = \theta_{gen}(s_i, k_i), \quad f_{i_{\ell}} = \theta_{gen}(s_i, d_{i_\ell}), \; \ell \in \{1, 2, 3\}.
\label{eq1}
\end{equation}
\textbf{Feature Extraction Module} 
The second module, $\theta_{ex}$, takes the stem, an option, and the corresponding reasoning steps or feedback message from Equation~\ref{eq1} as input. It generates a latent feature of the option. This module is based on finetuning an open-source LLM. Due to the large number of tokens in the reasoning steps and feedback messages, we use \textit{longformer-base-4096}, an encoder-only LLM designed to handle sequences of up to 4096 tokens. We denote the latent feature of the key and distractor as $h_{i_k}$ and $h_{i_\ell}$ respectively:
\begin{equation}
h_{i_k} = \theta_{ex}(s_i, k_i, r_i), \quad h_{i_\ell} = \theta_{ex}(s_i, d_{i_\ell}, f_{i_{\ell}}), \; \ell \in \{1, 2, 3\}.
\label{eq2}
\end{equation}
\textbf{Student Interaction Module}
Inspired by IRT, which conceptualizes student knowledge levels as samples drawn from a standard normal distribution, we adopt a similar approach: sample $M$ student knowledge levels from the multivariate standard normal distribution, denoted as $L$.
$L$ represents a comprehensive set of student knowledge levels that encompass the overall abilities of a diverse group of students. It is important to note that $L$ remains fixed throughout the entire learning process. Although our dataset includes student response data, we leave the task of simultaneously learning a student 
knowledge level distribution in future work. This module, $\theta_{si}$,  
takes the latent feature of all options in an MCQ and $L$ as input. It predicts the likelihood of students, on average, selecting each option of an MCQ. We use a 2-layer MLP with leaky\_relu as the activation function for this module. Specifically, For a given student $j$, the dimensionality of the latent feature of each option $h_i$ from Equation~\ref{eq2} is not compatible with the dimensionality of the student’s knowledge level $L_j$. To address this mismatch, we project $L_j$ into the same representation space as $h_i$ using $\theta_{si}$. Let the projected student knowledge level be denoted as $T_j$:
\begin{equation}
T_j = \theta_{si}(L_j).
\label{eq3}
\end{equation}
We calculate the option selection likelihood distribution of student $j$ by computing the dot product between $h_i$ and $T_j$ from Equation~\ref{eq3}. The resulting values are normalized using the Softmax function. Let the selection likelihoods for the key and distractors be denoted as $p_{j_k}$ and $p_{j_\ell} (\ell \in \{1, 2, 3\})$, respectively:  
\begin{equation}
\small
p_{j_k} = \frac{\exp(h_{i_k}^\top T_j)}{\exp(h_{i_k}^\top T_j) + \sum_{\ell=1}^3 \exp(h_{i_\ell}^\top T_j)}, \quad
p_{j_\ell} = \frac{\exp(h_{i_\ell}^\top T_j)}{\exp(h_{i_k}^\top T_j) + \sum_{\ell'=1}^3 \exp(h_{i_{\ell'}}^\top T_j)}.
\label{eq4}
\end{equation}
\textbf{Difficulty Prediction Module}
The last module, $\theta_{dp}$, takes the predicted likelihood of all $M$ students selecting each option for $Q_i$ from Equation~\ref{eq4} as input, and predicts the difficulty of $Q_i$. We use a 4-layer MLP with tanh as the activation function for this module. We denote the predicted difficulty as $\hat{df}_i$:
\begin{equation}
\hat{df}_i = \theta_{dp}(\textstyle \sum_{j=1}^M [p_{j_k}, p_{j_1}, p_{j_2}, p_{j_3}])/M.
\label{eq5}
\end{equation}

\subsection{Method Intuition}
Instead of directly predicting MCQ difficulty, our method adopts a two-stage approach. First, it predicts the likelihood that average students will select each option. Second, it uses these likelihoods to predict the difficulty of the MCQ. The intuition behind our method is that the likelihood of average students selecting each option provides valuable insights into MCQ difficulty. It captures both the complexity of reaching the key and the plausibility of distractors from student perspective.
If the likelihood of selecting the key is significantly higher than that of the distractors, the question is likely easy because most students can identify the correct answer without much confusion. On the other hand, if the selection distribution is more evenly distributed among the key and distractors, it suggests that the distractors are highly plausible and contain misconceptions or errors that real students are likely to have. This confusion leads to fewer students selecting the key, which corresponds to a higher difficulty level. 

We ensure the accuracy of predicted average student selection
likelihood distribution from two perspectives. From the student perspective, we sample a comprehensive set of student knowledge levels from a multivariate standard normal distribution. Unlike the one-dimensional representation used in IRT, which condenses student knowledge level to a single value, a multi-dimensional representation considers multiple skills separately. This representation captures the variations in student proficiency across different skills, providing a more accurate reflection of their overall knowledge level. By incorporating this comprehensive set of student knowledge levels, the predicted distribution is unbiased.


From the MCQ perspective, predicting accurate student selection likelihood distribution using only the stem and options is challenging. The reason is that these components fail to convey the thought process required to reach the key and explain the errors students make to select distractors explicitly. To address this challenge, we incorporate auxiliary information in the form of reasoning steps and feedback messages. The reasoning steps outline the logical process needed to arrive at the key, while the feedback messages explain the potential errors or misconceptions that could lead students to select each distractor. This auxiliary information enhances the Feature Extraction Module's ability to capture the complexity of reaching the key and the plausibility of distractors, ultimately improving the accuracy of predicted student selection likelihood distribution.

\subsection{Training Process}
\begin{figure*}[t]
\centering
\includegraphics[width=1\linewidth]
{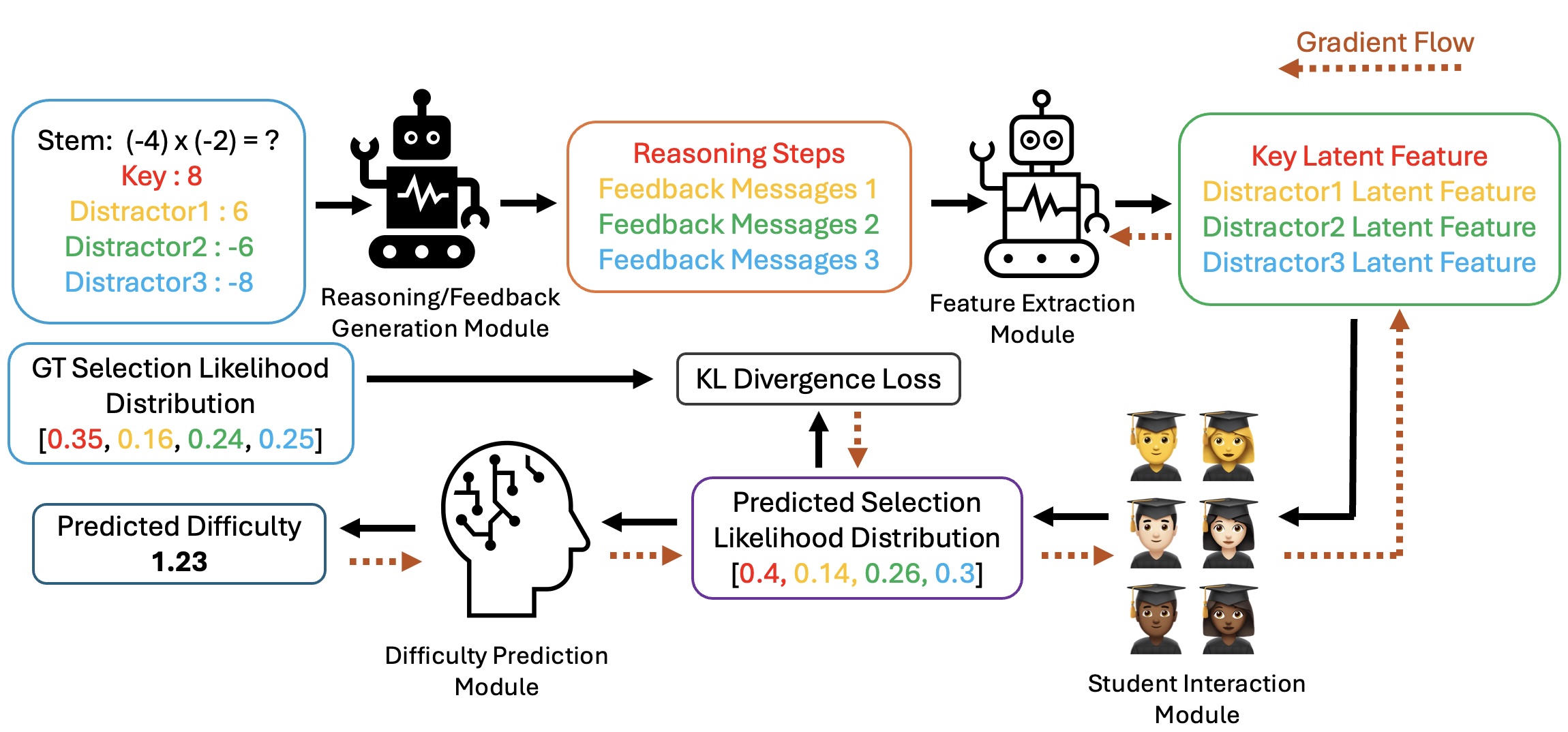}
\caption{Overview of our MCQ difficulty prediction pipeline.
}
\label{fig:training_process}
\vspace{-0.5cm}
\end{figure*}
 
The difficulty prediction loss for $Q_i$ is defined as the negative log-likelihood of the ground truth difficulty $df_i$ given the predicted difficulty from Equation~\ref{eq5}:
\begin{equation*}
\mathcal{L}_{d_i} = -\log P(df_i | \hat{df}_i).
\end{equation*}
To further encourage our method to predict the average student selection likelihood distribution accurately, we calculate the ground truth distribution by normalizing $[n_{k_i}, n_{d_{i_1}}, n_{d_{i_2}}, n_{d_{i_3}}]$, i.e., the number of students selecting each option. We then introduce a KL divergence loss between the predicted and ground truth distributions. We denote the KL divergence loss as $\mathcal{L}_{kl_i}$. The final training objective is to minimize:
\begin{equation*}
\mathcal{L}_{i} = \mathcal{L}_{d_i} + \alpha * \mathcal{L}_{kl_i},
\end{equation*}
where $\alpha$ is a hyperparameter that controls the relative weight of the KL divergence loss. See Figure~\ref{fig:training_process} for how gradient flows from loss to modules.
For all $N$ MCQs, we train the method end-to-end by optimizing its parameters to minimize $\mathcal{L}$:
\begin{equation*}
\mathcal{L} = (\textstyle \sum_{i=1}^N \mathcal{L}_{i})/N.
\end{equation*}

\vspace{-0.6cm}

\section{Experiments}
In this section, we detail the two real-world datasets used in our experiments, outline the pre-processing steps, define the baselines and evaluation metrics, detail the experimental setup, and analyze the quantitative and qualitative results.

\subsection{Datasets and Pre-processing steps}
We use two real-world datasets containing math MCQs and corresponding student response data. The first dataset originates from an adult proficiency math test administered by a US state.
The dataset contains 517 math MCQs. We manually review each MCQ and identify 191 math MCQs containing diagrams and set them aside for future studies, resulting in 317 math MCQs in the final dataset. This dataset includes IRT difficulty parameters calibrated using the 3PL-IRT model on 2024 student response data, serving as ground truth difficulties. The range of the ground truth difficulty is from -1.614 to 2.487. Additionally, we analyze option selection counts from students spanning 2008 to 2024, aggregating over years within this period. We denote this dataset as \textbf{APT}. 

The second dataset originates from the NeurIPS 2020 Education Challenge, provided by Eedi, a digital learning platform in the UK \cite{wang2021educational}. It contains 948 math MCQs, with all the components of the math MCQ provided as images. We extract the math MCQ components using an optical character recognition tool and manually review each math MCQ to ensure all components are accurately captured. During this process, we identify 621 math MCQs containing diagrams and set them aside for future studies, resulting in 327 math MCQs in the final dataset. This dataset includes 516,567 responses from 5,528 students.
Since the dataset does not include IRT parameter estimates, we estimate them on student responses using the 2PL-IRT model, and use the resulting difficulty estimates as the ground truth difficulty. The range of the ground truth difficulty is from -1.862 to 3.0. We denote this dataset as \textbf{EEDI}.

\subsection{Baselines and Evaluation Metrics}
For the baselines, we use three methods ranging from simple linear regression to finetuning LLMs. Inspired by prior work \cite{srivatsa2024makes}, the first baseline involves extracting 9 syntactic features from the math MCQs using NLP parsers. These features are divided into two categories: textual and mathematical. The 5 textual features include the number of sentences, the number of nouns, the number of unique nouns, the number of prepositions, and the Flesch-Kincaid readability score. The 4 mathematical features include the number of numerical values, the number of text-based numerical values, the number of operators, and the number of unique operators. We use these 9 features as input and train a linear regression model to predict the difficulty of the math MCQ by using the mean squared error loss to minimize the difference between predicted and ground truth difficulties. We denote this baseline as \textbf{LR}. 

Inspired by prior work \cite{benedetto2021application}, we also finetune an encoder-only LLM to predict the math MCQ difficulty by leveraging its well-known ability to extract latent features from the text. The LLM takes the tokenized text of the math MCQ, including the question stem and options with this order, $[s_i, k_i, d_{i_{1}}, d_{i_{2}}, d_{i_{3}}]$, and uses the [CLS] token embedding to represent the entire sequence. A linear regression head is added on top of the [CLS] token embedding to predict the difficulty. For a fair comparison with our method, we use \textit{longformer-base-4096} as the LLM. We denote this baseline as \textbf{FT}.


Moreover, building on our observation that reasoning steps for the key and feedback messages for the distractors enhance the LLM’s understanding of the complexity of reaching the key and the plausibility of the distractors, we include a strong baseline that incorporates this auxiliary information. Specifically, we append the reasoning steps before the key and feedback message before each distractor as the input, $[s_i, r_i, k_i, f_{i_{1}}, d_{i_{1}}, f_{i_{2}}, d_{i_{2}}, f_{i_{3}}, d_{i_{3}}  ]$. For a fair comparison with our method, we use \textit{longformer-base-4096} as the LLM. We denote this enhanced baseline as FT with reasoning, or \textbf{FTWR}. 


To evaluate the performance of math MCQ difficulty prediction methods, we use two absolute metrics: Mean Squared Error (\textbf{MSE}) and the Coefficient of Determination (\bm{$R^2$}) \cite{chicco2021coefficient}. MSE measures the average squared difference between the predicted and ground truth difficulties, with lower values indicating better predictive accuracy. $R^2$ measures the proportion of variance in the ground truth difficulty that is explained by the method’s predictions; values closer to 1 indicate better predictive accuracy, while negative values indicate performance worse than simply predicting the mean of the ground truth difficulty.

Moreover, we use a new ranking metric to evaluate the methods’ ability to capture the difficulty ranking between math MCQs. For each pair of math MCQs, we compare predicted difficulties to determine their relative ranking, i.e., whether the first math MCQ is predicted to have higher or lower difficulty than the second one. The metric is calculated as the percentage of math MCQ pairs where the predicted ranking matches the ground truth ranking; values closer to 1 indicate a higher accuracy in preserving the correct relative difficulty ranking between pairs of math MCQs. We denote this metric as \textbf{MATCH}.

\subsection{Experimental Setup}
 To enable efficient training, we pre-sample a comprehensive set of student knowledge levels. The same set of student knowledge levels is consistently used across all math MCQs for training, validation, and testing. Based on our experiment results, we find that sampling $M=1,000$ student knowledge levels from a d-dimensional standard normal distribution, with d = 2, provides the best performance.
In our experiments, We split the data into training, validation, and test sets in a ratio of 6.5:1.5:2, and the final method weights are obtained based on the minimum validation loss. To ensure robust and reliable results, especially given the limited number of math MCQs in the datasets, we perform 5-fold cross-validation across all experiments. We train all methods on one NVIDIA RTX A6000 GPU. For our method, we use a batch size of 10 and train for 300 epochs with learning rates set to 1e-5 for the Feature Extraction Module, 1e-2 for the Student Interaction Module, and 1e-2 for the Difficulty Prediction Module. The hyperparameter  $\alpha$  is set to 0.0886. For LR, we calculate the closed-form solution. For FT and FTWR, we use a batch size of 16 and train for 30 epochs with a learning rate of 1e-5. For prompting \textit{GPT-4o} via the OpenAI API, we set the temperature to 1.0, the max-tokens to 1000, and the top-p to 0.95.


\subsection{Quantitative Results}
\begin{table}[t]
\caption{Results of math MCQ difficulty prediction task on absolute and ranking metrics across both datasets. The results demonstrate that our method outperforms the baselines in most of the cases.}
\vspace{10pt}
\centering
\renewcommand{\arraystretch}{1.5} 
\setlength{\tabcolsep}{6pt} 
\label{tab:math_text_tokens}
\begin{tabular}{|c|c|c|c|c|}
\hline
\textbf{Data}     & \textbf{Method} & \textbf{MSE}         & \bm{$R^2$}        & \textbf{MATCH}       \\ \hline
\multirow{4}{*}{APT} 
                      & Our method       & \textbf{0.521 ± 0.059} & \textbf{0.536 ± 0.043} & 0.770 ± 0.021          \\ \cline{2-5} 
                      & FTWR                & 0.566 ± 0.079          & 0.495 ± 0.070          & \textbf{0.780 ± 0.027} \\ \cline{2-5} 
                      & FT                  & 0.622 ± 0.090          & 0.448 ± 0.063          & 0.756 ± 0.009          \\ \cline{2-5} 
                      & LR                  & 0.984 ± 0.124          & 0.125 ± 0.100          & 0.641 ± 0.026          \\ \hline
\noalign{\vskip 2pt} 
\hline
\multirow{4}{*}{EEDI} 
                      & Our method        & \textbf{0.367 ± 0.082} & \textbf{0.525 ± 0.101} & \textbf{0.757 ± 0.032} \\ \cline{2-5} 
                      & FTWR                & 0.471 ± 0.149          & 0.390 ± 0.181          & 0.719 ± 0.056          \\ \cline{2-5} 
                      & FT                  & 0.522 ± 0.079          & 0.329 ± 0.048          & 0.714 ± 0.019          \\ \cline{2-5} 
                      & LR                  & 0.688 ± 0.028          & 0.084 ± 0.059          & 0.599 ± 0.036          \\ \hline
\end{tabular}
\label{tab:result}
\vspace{-5pt}
\end{table}

Table~\ref{tab:result} shows the performance of our method and three baselines on all metrics across the APT and EEDI datasets. We observe that our method outperforms all baselines in both MSE and $R^2$ metrics across two datasets. Notably, Especially on the EEDI dataset, our method achieves a 28.3\% reduction in MSE compared to the best baseline, FTWR. It also improves $R^2$ by 34.6\%. This improvement indicates that our method explains a larger proportion of the variance in math MCQ ground truth difficulty. These results suggest that predicting the average student selection likelihood distribution before predicting difficulty offers additional benefits. This intermediate step explicitly learns how students with varying knowledge levels interact with math MCQs. By first learning this interaction, the method captures both the complexity of reaching the key and the plausibility of distractors from student perspectives, which helps to effectively distinguish between easy and challenging math MCQs. In contrast, methods that directly predict difficulty, such as FTWR, do not capture how different factors contribute to student decision-making. For the MATCH metric, the difference between our method and FTWR is relatively small on the EEDI dataset, while FTWR achieves a higher score on the APT dataset. This result suggests that both methods effectively learn the relative ranking of math MCQ difficulties. 

FTWR outperforms FT across all metrics on both datasets. These results highlight the importance of incorporating reasoning steps and feedback messages in predicting math MCQ difficulty. These auxiliary inputs provide additional context on the complexity of reaching the key and the plausibility of distractors, thus enabling the LLM to capture the difficulty of the math MCQ better. In contrast, FT, which relies solely on the question stem and options, lacks this enriched information, leading to less accurate difficulty predictions. 

FT outperforms LR across all metrics on both datasets. These results show the advantage of using a pre-trained LLM to capture rich contextual information within math MCQs. In contrast, LR, which relies on handcrafted syntactic features, lacks the ability to understand the relationships between the question stem and options, leading to less accurate difficulty predictions.

\vspace{10pt}
\begin{figure}[b]
    \centering
    \includegraphics[width=0.47\textwidth]{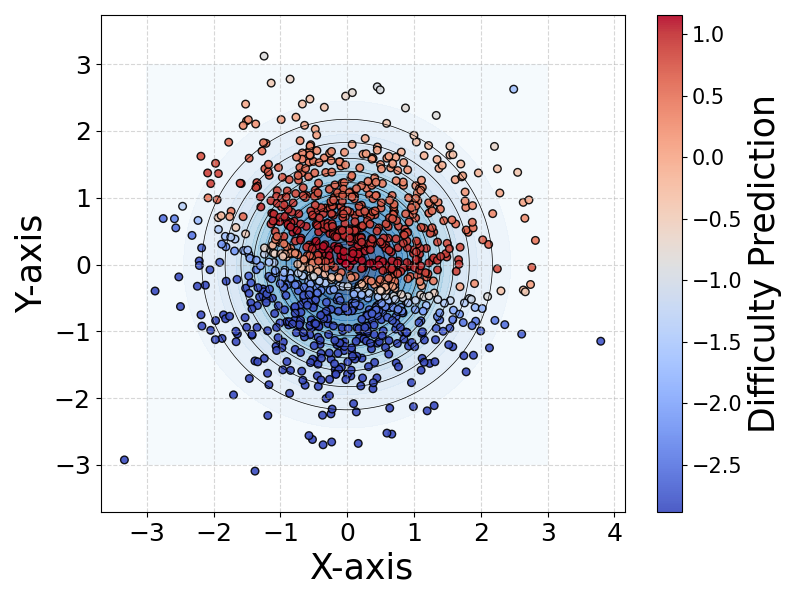}
\includegraphics[width=0.52\textwidth,height=1.5\textwidth,keepaspectratio]{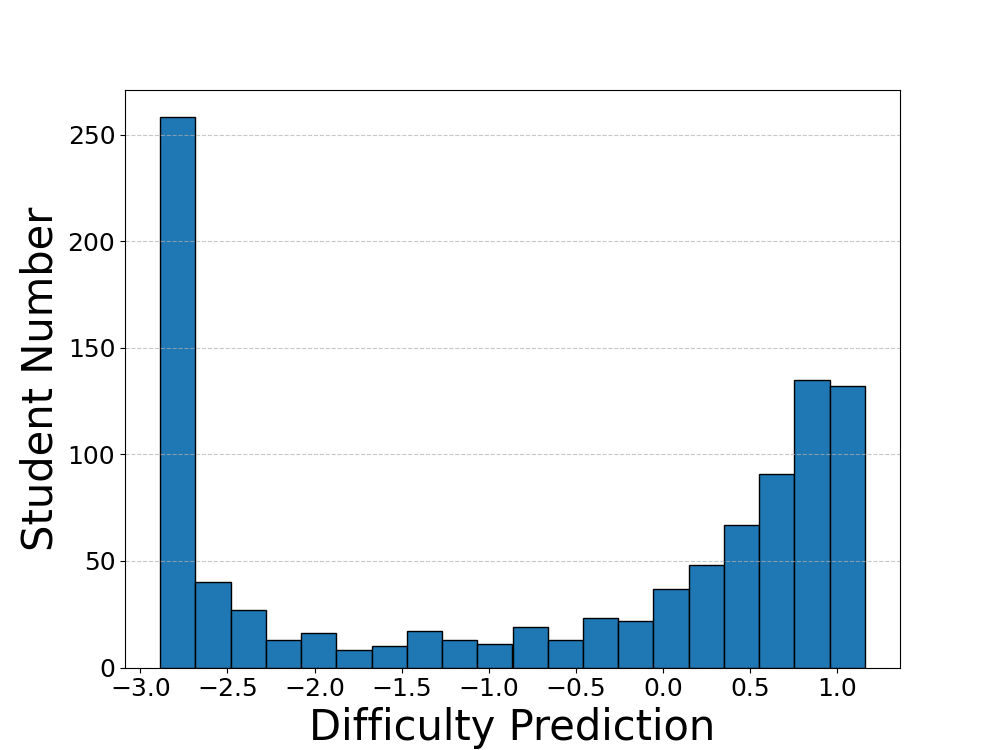}
    \caption{Sampled student knowledge (left) and question difficulty (right).}
    \label{fig:student_distribution}
\end{figure}

\subsection{Qualitative Analysis}
\begin{table}[h]
\caption{An example MCQ showing how reasoning steps and feedback messages improve the accuracy of difficulty prediction.}
\vspace{10pt}
\centering
\renewcommand{\arraystretch}{1.2} 
\scalebox{.95}{
\begin{tabular}{|m{12.75cm}|}
    \hline
    \textbf{Solve this equation: (-11) + 7 = ?} \\
    \textbf{Key}: \textbf{-4} \hspace{30pt} \textbf{Distractor1}: \textbf{18}\hspace{30pt} \textbf{Distractor2}: \textbf{-18}\hspace{30pt} \textbf{Distractor3}: \textbf{-5} \\
    \hline
    \textbf{Reasoning Steps}: To find the number that belongs in the blank, we need to evaluate the expression (-11) + 7. \\
    1. Start with the first number: -11 \\
    2. Add 7 to -11: (-11) + 7 = -4 \\
    Therefore, the number that belongs in the blank is -4. \\
    \textbf{Feedback Message1}: The student appears to have misunderstood the operation or the numbers involved. They likely added the absolute values of 11 and 7 (11 + 7 = 18) without considering the negative sign. This suggests a confusion between positive and negative numbers. \\
    \textbf{Feedback Message2}: The student subtracted 7 from 11 and kept the negative sign, resulting in -18. This indicates a misinterpretation of the order of operations or a misunderstanding of how to handle negative numbers. \\
    \textbf{Feedback Message3}: The student reached -5, which is close to the correct answer. They likely made a small arithmetic error when adding 7 to -11, possibly by miscalculating the result of -11 + 7. \\
    \hline
\end{tabular}}

\vspace{-1pt} 

\scalebox{.95}{
\begin{tabular}{|>{\centering\arraybackslash}p{2.2cm}|>{\centering\arraybackslash}p{3.2cm}|>{\centering\arraybackslash}p{3cm}|>{\centering\arraybackslash}p{3cm}|} 
    \hline
    \textbf{GT Difficulty} & \textbf{Our Method Prediction} & \textbf{FTWR Prediction} & \textbf{FT Prediction} \\
    \hline
    \textbf{-0.734} & \textbf{-0.387} & \textbf{-0.347} & \textbf{0.0161} \\
    \hline
\end{tabular}}
\label{tab:qua}
\end{table}

We now qualitatively investigate how different components of our method, such as reasoning steps, sampling from student knowledge level distribution, and explicit option selection likelihood modeling, help us to predict MCQ difficulty using the MCQs in EEDI dataset.

The example in Table~\ref{tab:qua} shows that, by incorporating reasoning steps for the key and feedback messages for the distractors as additional input to the LLM, FTWR predicts the difficulty of this MCQ more accurately than FT, which only uses the question stem and distractors as input. These reasoning steps (and feedback messages) explicitly give the LLM a sense of how difficult it is to reach the correct option and avoid incorrect options; otherwise, a small LLM, such as \textit{Longformer}, may not exhibit such reasoning capabilities. 

Figure~\ref{fig:student_distribution} provides some insights into the MCQ difficulty prediction task. The left part shows the samples we draw from the student knowledge distribution: the color of each point shows the predicted difficulty of the MCQ in Table~\ref{tab:qua} for one particular student. The right part shows a histogram plot of the predicted difficulties of the MCQ for all sampled student knowledge levels.
%
%
We see that both distributions are bi-polar, where most samples correspond to either very high or very low knowledge levels/difficulty, with a sharp transition boundary between the two regions and very few points in the middle. These patterns match, but are somewhat surprising, since the assumption in IRT is that most (ability) parameter values are centered around the mean. However, in practice, most real students answers either almost all questions correctly or very few, which may give rise to such bi-polar distributions. We leave a detailed examination of the learned parameters to future work. 
\section{Conclusions and Future Work}
In this work, we introduce a novel method for MCQ difficulty prediction. This method first augments the reasoning steps and feedback messages for each option, then samples student knowledge profiles from a distribution, and finally predicts the likelihood of each option being selected and use it to predict the MCQ's difficulty. Experiments on two real-world math MCQ response datasets show that our method consistently outperforms baselines. Avenues for future work includes 1) learning a student knowledge level distribution instead of sampling from a standard normal distribution, 2) explore predicting the discrimination parameter in 2PL IRT models, and 3) experiment on other subjects and domains beyond math, such as medical education and language learning.

\newpage
\bibliographystyle{splncs04}
\bibliography{main}
\end{document}